\documentclass[runningheads]{llncs}
\usepackage{threeparttable}
\usepackage{graphicx}
\usepackage{todonotes}
\usepackage{comment}
\usepackage{multirow}
\usepackage{booktabs}
\begin{document}
%
\title{Representation Learning for Person or Entity-Centric Knowledge Graphs: \\An Application in Healthcare} 


%
\titlerunning{Representation Learning for Person or Entity-centric Knowledge Graphs}

%
\author{Christos Theodoropoulos\thanks{Work completed while the first author was an intern at IBM Research Europe.}\inst{1, 2}
\and
Natasha Mulligan\inst{2}
\and
Thaddeus Stappenbeck\inst{3}
\and
Joao Bettencourt-Silva\inst{2}
}
\authorrunning{C. Theodoropoulos et al.}
%
\institute{KU Leuven, Oude Markt 13, 3000 Leuven, Belgium
\email{christos.theodoropoulos@kuleuven.be}, \and IBM Research Europe, Dublin, Ireland \\ \email{\{jbettencourt, natasha.mulligan\}@ie.ibm.com}, \and Lerner Research Institute, Cleveland Clinic, Cleveland, Ohio, United States \\ \email{stappet@ccf.org}}





%

%
\maketitle              
\begin{abstract} 

\vspace{-10mm}

Knowledge graphs (KGs) are a popular way to organise information based on ontologies or schemas and have been used across a variety of scenarios from search to recommendation. Despite advances in KGs, representing knowledge remains a non-trivial task across industries and it is especially challenging in the biomedical and healthcare domains due to complex interdependent relations between entities, heterogeneity, lack of standardization, and sparseness of data.

KGs are used to discover diagnoses or prioritize genes relevant to disease, but they often rely on schemas that are not centred around a node or entity of interest, such as a person. Entity-centric KGs are relatively unexplored but hold promise in representing important facets connected to a central node and unlocking downstream tasks beyond graph traversal and reasoning, such as generating graph embeddings and training graph neural networks for a wide range of predictive tasks.

This paper presents an end-to-end representation learning framework to extract entity-centric KGs from structured and unstructured data. We introduce a star-shaped ontology to represent the multiple facets of a person and use it to guide KG creation. Compact representations of the graphs are created leveraging graph neural networks and experiments are conducted using different levels of heterogeneity or explicitness. A readmission prediction task is used to evaluate the results of the proposed framework, showing a stable system, robust to missing data, that outperforms a range of baseline machine learning classifiers. We highlight that this approach has several potential applications across domains and is open-sourced. Lastly, we discuss lessons learned, challenges, and next steps for the adoption of the framework in practice. 

\vspace{-2mm}


\end{abstract}
%
%
%


\section{Introduction}

\vspace{-3mm}

Knowledge graphs (KGs) have been widely used to organize information in a structured and flexible way enabling a variety of downstream tasks and applications \cite{hogan2021knowledge}. KGs consist of nodes (entities) and edges (relations) between them, that represent the information in a particular domain or set of domains. The ability of KGs to support complex reasoning and inference has been explored in a variety of tasks including search \cite{noy2019industry,guha2003semantic,kasneci2008naga}, recommendation \cite{zhang2016collaborative,wu2022graph,guo2020survey}, and knowledge discovery \cite{nentwig2017survey,kumar2020link}. 

KGs are becoming increasingly used across a wide range of biomedical and healthcare applications. Knowledge graphs typically rely on information retrieved from biomedical literature and early approaches, such as BioGrid \cite{biogrid2012}, have first depended on manual curation of knowledge bases to map protein-to-protein interactions using genes and proteins as entity types. Semi-automatic and machine-learning approaches have been introduced to assist, for example, in finding associations or relations between important concepts such as diseases and symptoms \cite{relclassEMNLP2022}. Recently, healthcare and clinical applications have otherwise focused on building KGs from electronic health records (EHRs) where nodes represent diseases, drugs, or patients, and edges represent their relations \cite{KGsBiomedicine2020}. Most approaches are however limited by the challenging nature of healthcare data, including heterogeneity, sparseness and inconsistent or lacking standardization \cite{jbHeterogeneity2012}. Recent calls for future work on KGs have expressed the need to develop new models and algorithms that are able to take these challenges (e.g. missing data) into account \cite{KGsBiomedicine2020}. Furthermore, there is a need to be able to accurately represent information from multiple data sources about individual patients. This is not only required by physicians to support routine hospital activities but also for clinical research in, for example, developing novel predictive models or discovering new important features associated with poor patient outcomes. A holistic representation of individual patients should therefore capture not only their clinical attributes such as diagnoses, procedures, and medication but also other variables that may also be predictors such as demographics, behavioral and social aspects (e.g. smoking habits or unemployment). Incorporating new types of information in models and analyses will allow the creation of better tools for evaluating the effectiveness of therapies and directing them to the most relevant patients. 

This paper presents an end-to-end representation learning framework to extract information and organize it into entity-centric KGs from both structured and unstructured data. A star-shaped ontology is designed and used for the purpose of representing the multiple facets of a person and guiding the first stages of a person knowledge graph (PKG) creation. Graphs are then extracted, and compact representations are generated leveraging graph neural networks (GNNs). We evaluate our approach using a real-world hospital intensive care unit (ICU) dataset and a hospital readmission prediction task. To the best of our knowledge, the novelty of the proposed approach can be summarised as:

\vspace{-3mm}
\begin{itemize}

\item the first end-to-end framework for PKG extraction in Resource Description Framework (RDF) and PyTorch Geometric applicable format, using structured EHRs as well as unstructured clinical notes.
    
\item the first use of a star-shaped Health \& Social Person-centric Ontology (HSPO) \cite{hspo} to model a comprehensive view of the patient, focused on multiple facets (e.g. clinical, demographic, behavioral, and social).
    
\item a representation learning approach that embeds personal knowledge graphs (PKGs) using GNNs and tackles the task of ICU readmission prediction using a real-world ICU dataset.

\item the implementation proposed in this paper is open-sourced\footnote{https://github.com/IBM/hspo-ontology}, adaptable, and generalizable so that it can be used to undertake other downstream tasks.

\end{itemize}

\vspace{-2mm}

The paper is structured in the following way: section 2 discusses related work and  section 3 includes a description of the HSPO ontology. Section 4 describes the data preprocessing pipeline for PKG extraction using the ontology and the processed dataset. Section 5 evaluates different graph approaches in tackling downstream tasks anchored by an ICU hospital readmission prediction task. Finally, section 6 discusses the impact and adoption of the study and highlights the applicability of the framework in different downstream tasks and domains. 

\vspace{-5mm}


\section{Related Work}

\vspace{-3mm}

Public KGs are perhaps the most pervasive type of knowledge graphs today. These are often based on publicly available documents, encyclopedias or domain-specific databases and their schemas describe the features typically found within them. In recent years, especially in the health and biomedical domains, different types of KGs have been proposed from literature or EHRs yet they are not usually centred around the individual. Recent works include the PubMed KG \cite{pmkg2020}, enabling connections among bio-entities, authors, articles, affiliations, and funding, the clinical trials KG \cite{ctkg2022}, representing medical entities found in clinical trials with downstream tasks including drug-repurposing and similarity searches, and the PrimeKG \cite{primeKG2023}, a multimodal KG for precision medicine analyses centred around diseases. Indeed disease-centric KGs have been previously proposed and despite some efforts in overlaying individual patient information \cite{nelson2019integrating}, these graphs are not centred around the person or patient.

The idea behind entity-centric knowledge graphs and particularly person-centric graphs is relatively new and unexplored. One of the first efforts to define personal knowledge graphs (PKGs) was that of Balog and Kenter \cite{balog2019personal} where the graph has a particular “spiderweb” or star-shaped layout and every node has to have a direct or indirect connection to the user (i.e. person). The main advantage of having such a star-shaped representation lies in the fact the graph itself becomes personalized, enabling downstream applications across health and wellbeing, personal assistants, or conversational systems \cite{balog2019personal}. Similarly, a review paper \cite{rastogi2020personal} discussed the idea of a Person Health KG as a way to represent aggregated multi-modal data including all the relevant health-related personal data of a patient in the form of a structured graph but several challenges remained and implementations are lacking especially in representation learning. Subsequent works have proposed personal research knowledge graphs (PRKGs) \cite{prkg_acm2022} to represent information about the research activities of a researcher, and personal attribute knowledge bases \cite{papc_acm2022} where a pre-trained language model with a noise-robust loss function aims to predict personal attributes from conversations without the need for labeled utterances. A knowledge model for capturing dietary preferences and personal context \cite{phkg_arxiv2021} has been proposed for personalized dietary recommendations where an ontology was developed for capturing lifestyle behaviors related to food consumption. A platform that leverages the Linked Open Data \cite{bizer2011linked} stack (RDF, URIs, and SPARQL) to build RDF representations of Personal Health Libraries (PHLs) for patients is introduced in \cite{ammar2021using} and is aimed at empowering care providers in making more informed decisions. A method that leverages personal information to build a knowledge graph to improve suicidal ideation detection on social media is introduced by \cite{cao2020building}. In the latter, the extracted KG includes several user nodes as the social interaction between the users is modeled. UniSKGRep \cite{shen2023uniskgrep} is a unified representation learning framework of knowledge graphs and social networks. Personal information of famous athletes and scientists is extracted to create two different universal knowledge graphs. The framework is used for node classification \cite{xiao2022graph,rongdropedge} and link prediction \cite{kumar2020link,martinez2016survey,lu2011link}. A graph-based approach that leverages the interconnected structure of personal web information, and incorporates efficient techniques to update the representations as new data are added is proposed by \cite{safavi2020toward}. The approach captures personal activity-based information and supports the task of email recipient prediction \cite{qadir2016activity}.

Despite the above efforts and idiosyncrasies, person-centric knowledge graphs have not been extensively used for predictive or classification tasks, especially those involving graph embeddings and GNNs. Similarly, ontologies that support the creation of entity-centric, star-shaped PKGs are not well established and there is no published research on representation learning of person-centric knowledge graphs using GNNs. 
To the best of our knowledge, this paper is the first to propose a framework for learning effective representations of an entity or person (i.e., graph classification setting \cite{zhang2018end,cai2018comprehensive,wu2020comprehensive}) using PKGs and GNNs that can be applied to different downstream predictive tasks.

\vspace{-4mm}

\section{HSPO Ontology}

\vspace{-3mm}

The Health and Social Person-centric ontology (HSPO) has been designed to describe a holistic view of an individual spanning across multiple domains or facets. The HSPO defines a schema for a star-shaped Personal Knowledge Graph \cite{balog2019personal} with a person as a central node of the graph and corresponding characteristics or features (e.g. computable phenotype) linked to the central node.


This view is unique as it is designed to continue to be expanded with additional domains and facets of interest to the individual including but not limited to the clinical, demographics and social factors in a first version but also being able to expand future versions to include behavioral, biological, or gene information. Representing a holistic view of individuals with new information generated beyond the traditional healthcare setting is expected to unlock new insights that can transform the way in which patients are treated or services delivered.

Previous ontologies have been built to harmonize disease definitions globally (MONDO \cite{MONDO}), to provide descriptions of experimental variables such as compounds, anatomy, diseases, and traits (EFO \cite{EFO}), or to describe evidence of interventions in populations (\cite{PICO}). Other ontologies have focused on specific contexts or diseases, such as SOHO, describing a set of social determinants affecting individual’s health \cite{SOHO} or an ontology that describes behavior change interventions \cite{HBCP}. Further to this, not all ontologies provide mappings for their entities into standard biomedical terminologies and vocabularies. The HSPO aims to address these challenges by being the first to create a person-centric view linking multiple facets together, leveraging existing ontological efforts, and providing mappings to known terms of biomedical vocabularies when appropriate.

The HSPO ontology has been built incrementally with the main objective of providing an accurate and clear representation of a person and their characteristics across multiple domains of interest. Domains of interest were identified and prioritized together with domain experts and a well-established methodology \cite{ontology101} was followed to guide development. The HSPO is built not only to ensure that questions may be asked from the generated KGs but also that derived graphs may be used to train neural networks for downstream tasks.

\vspace{-7mm}
\begin{figure}[!h]
  \centering
  \includegraphics[scale=0.3]{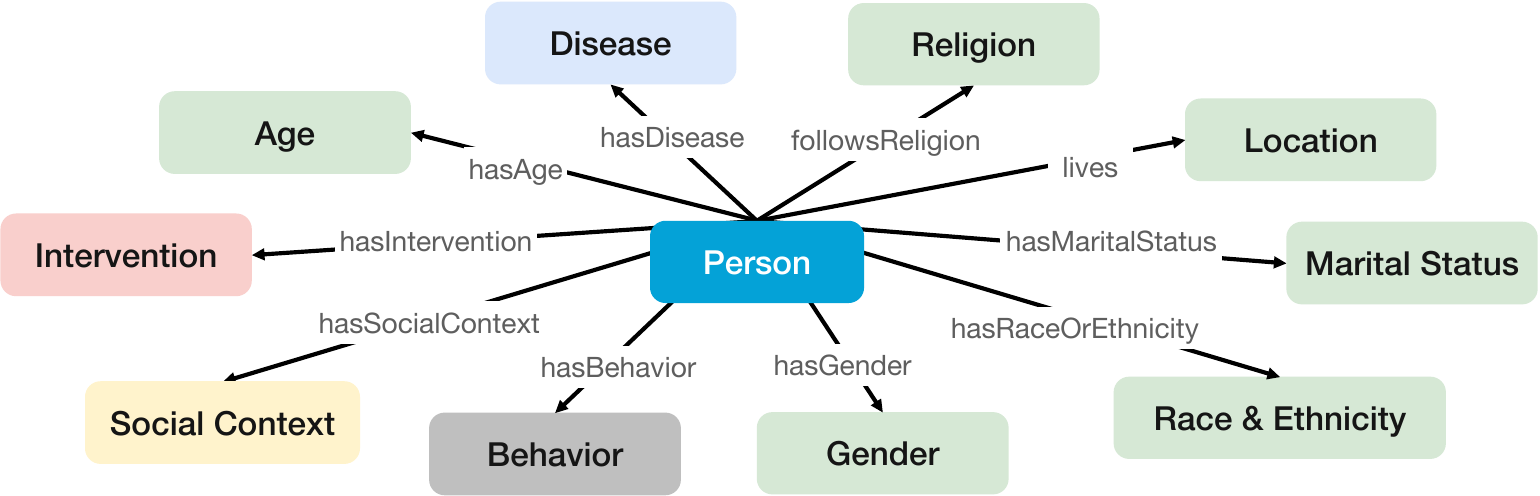}
  \vspace{-4mm}
  \caption{Main classes defined in the HSPO.}
\end{figure}
\vspace{-14mm}

\section{Person-Centric Knowledge Graph Extraction} 

\vspace{-3mm}

Generating Person-Centric Knowledge Graphs requires a data preprocessing pipeline to prepare the dataset before graphs can be extracted. This section describes the use case rationale and data, the data preprocessing pipeline, and the steps taken to extract knowledge graphs. 

\vspace{-3.5mm}

\subsection{Use Case Rationale and Dataset}

\vspace{-2.5mm}

In this study, we use EHRs on ICU admissions provided by the MIMIC-III dataset \cite{johnson2016mimic,johnson2016mimic2,goldberger2000physiobank}, a well-established publicly available and de-identified dataset that contains hospital records from a large hospital’s intensive care units. MIMIC-III, therefore, includes data and a structure similar to most hospitals, containing both tabular data and clinical notes in the form of free text. More precisely, the data covers the demographic (e.g. marital status, ethnicity, etc.), clinical (e.g. diagnoses, procedures, medication, etc.), and some aspects of the social view of the patient embedded in text notes. Detailed results of lab tests and metrics of monitoring medical devices are also provided.

MIMIC-III is not only an appropriate dataset because of its structure and the types of data that it contains but also because of its population characteristics, including the reasons for hospital admission. More than 35\% of the patients in this dataset were admitted with cardiovascular conditions \cite{johnson2016mimic2} which are broadly relevant across healthcare systems globally. Indeed, following their first hospital discharge, nearly 1 in 4 heart failure patients are known to be readmitted within 30 days, and approximately half are readmitted within 6 months \cite{khan2021trends}. These potentially avoidable subsequent hospitalizations are on the increase and reducing 30-day readmissions has now been a longstanding target for governments worldwide to improve the quality of care (e.g. outcomes) and reduce costs \cite{lawson2021trends}.

Therefore, both the experiments carried out and knowledge graphs generated in this paper describe a use case on patients admitted with cardiovascular conditions and a downstream prediction task to identify potential 30-day readmissions. This task can be further generalised to other conditions as readmissions are a widely used metric of success in healthcare, and other outcome metrics (e.g. specific events, mortality) are also possible using the approach proposed in this paper and accompanying open-source code repository.

\vspace{-5mm}

\subsection{Preprocessing Pipeline}

\vspace{-10mm}

\begin{figure}[!h]
  \centering
  \includegraphics[scale=0.75]{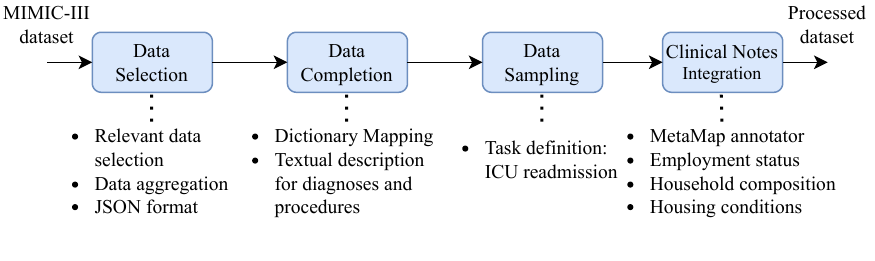}
  \vspace{-9mm}
   \caption{Data preprocessing pipeline: The EHRs of MIMIC-III are being preprocessed and a unified JSON file is extracted, that consists of records for each admission with the essential information about the clinical, demographic, and social view of the patient.}
\end{figure}

\vspace{-13.5mm}

\subsubsection{Data Selection and Completion}

The goal of the data preprocessing is to prepare the dataset for the PKG extraction. Due to the nature of MIMIC-III, each PKG represents the state of a patient during a single hospital admission. In the first step of the data preprocessing pipeline (Fig. 2), we select and aggregate the relevant data. In order to construct an efficient and applicable representation for a range of downstream tasks, the data selection step is necessary to include a subset of the EHR data, that is concise.
We exclude the detailed lab test results (e.g. full blood count test values), as we assume that the diagnoses, medication, and procedures data are sufficiently expressive to represent the clinical view of the patient. The inclusion of fine-grain information poses additional challenges in the encoding of the PKG and the representation learning process.  

Following this strategy, we select the demographic information: gender, age, marital status, religion, ethnicity, and the essential clinical data: diagnoses, medication, and procedures. We create a different record for each admission with the corresponding information using the JSON format. The diagnoses and procedures in the MIMIC-III dataset are recorded using ICD-9 \cite{world1980international,world1998icd} (The International Classification of Diseases, Ninth Revision) coding schema with a hierarchical structure (with a format \textit{xXXX.YYY}) where the first three or four digits of the code represent the general category, and the subsequent three digits after the dot separator represent more specific subcategories. 

We group the diagnoses and procedures using the corresponding general family category (\textit{xXXX}) to reduce the number of different textual descriptions while defining substantially each diagnosis and procedure. For example, the diagnoses \textit{acute myocardial infarction of anterolateral wall} (ICD-9 code: 410.0) and \textit{acute myocardial infarction of inferolateral wall} (ICD-9 code: 410.2) are grouped under the general family diagnosis \textit{acute myocardial infarction} (ICD-9 code: 410). 

This grouping is important, otherwise, the encoding of the graph and training of a Graph Neural Network to solve a downstream task would be very challenging, as some diagnoses and procedures are rare and underrepresented in the limited dataset of the study. In detail, more than 3,900 diagnoses and 1,000 procedures have a frequency of less than 10 in the dataset, while after grouping the numbers drop to less than 300 and 280 respectively. 

\vspace{-5mm}

\subsubsection{Data Sampling}

As the diagnoses and procedures are given using ICD-9 coding, we add the textual descriptions of the ICD-9 codes to the dataset. We sample the data records that are appropriate for the selected downstream task of the paper: 30-day ICU readmission prediction. The day span is a hyperparameter of the approach. The ICU admission records of patients that passed away during their admission to the hospital, or in a day span of fewer than 30 days after their discharge, are excluded. 

\vspace{-5mm}

\subsubsection{Clinical Notes Integration}

The clinical notes of the dataset contain information related to the clinical, demographic, and social aspects of the patients. We use the MetaMap \cite{aronson2001effective} annotator to annotate the clinical notes with UMLS entities (UMLS Metathesaurus \cite{bodenreider2004unified,mccray2003upper}) and sample the codes that are related to certain social aspects such as employment status, household composition, and housing conditions. In this study, we focus on these three social aspects because social problems are known to be non-clinical predictors of poor outcomes but also they are often poorly recorded yet previous works have identified these specific three social problems available in MIMIC-III \cite{lybarger2021annotating}. The extracted UMLS annotations of the clinical notes are integrated into the processed dataset. 
\vspace{-5mm}

\subsubsection{Summary of Processed Dataset}

After the data preprocessing steps, the total number of admission records is 51,296. 
From these, in 47,863 (93.3\%) of the cases, the patients are not readmitted to the hospital, while 3,433 (6.7\%) of the patients returned to the hospital within 30 days. In the downstream task for this paper, we focus on patients diagnosed with heart failure or cardiac dysrhythmia. Hence, the final dataset consists of 1,428 (9.2\%) readmission cases and 14,113 (90.8\%) non-readmission cases.

Hence, the dataset is highly imbalanced as the readmission cases are underrepresented. In addition, EHR data is often incomplete \cite{wells2013strategies,hu2017strategies,beaulieu2018characterizing,stiglic2019challenges} and MIMIC-III, as it consists of real-world EHRs, is no exception. Information may be missing for multiple reasons, such as the unwillingness of the patient to share information, technical and coding errors, among several others \cite{jbHeterogeneity2012}. More precisely, we observe that for some fields, such as religion, marital status, and medication there is a significant percentage of missing information. Tab. 1 shows the number of missing records per field for all admissions. We highlight that the social information (employment, housing conditions, and household composition), extracted from the unstructured data is scarce. This is an indication that the clinical notes focus predominantly on the clinical view of the patients, without paying attention to aspects that can be connected to the social determinants of health, as previously reported \cite{sdohextract2020,lybarger2021annotating}.


\vspace{-5mm}
\begin{table}[!h]
    \vspace{-2.5mm}
    \caption{Missing information in the processed dataset.}
    \vspace{-2.5mm}
    \centering
    \begin{tabular}{|cc|}
        \hline
        \textbf{Information} & \textbf{Records with missing information}\\
        \hline
        Gender & 0 \\
        \hline
        Religion & 17,794 (34.69\%)\\
        \hline
        Marital Status & 9,627 (18.77\%)\\
        \hline
        Race/Ethnicity & 4,733 (9.23\%)\\
        \hline
        Diseases/Diagnoses & 10 (0.02\%)\\
        \hline
        Medication & 8,032 (15.66\%)\\
        \hline
        Procedures & 6,024 (11.74\%)\\
        \hline
        Employment & 25,530 (49.77\%)\\
        \hline
        Housing conditions & 49,739 (96.96\%)\\
        \hline
        Household composition & 42,958 (83.75\%)\\
        \hline
    \end{tabular}
    \vspace{-6mm}
\end{table}

The MIMIC-III dataset is protected under HIPPA regulations. Thus, the detailed data distribution per field (e.g. race, diseases, medication, etc.) cannot be shared publicly. The implementation for the extraction of the distributions is publicly available in the official repository accompanying this paper and can be used when access to the dataset is officially granted\footnote{https://physionet.org/content/mimiciii/1.4/} to the user.

\vspace{-6mm}

\subsection{Person-Centric Knowledge Graph Extraction}

\begin{figure}[!h]
  \vspace{-11.9mm}
  \centering
  \includegraphics[scale = 0.1]{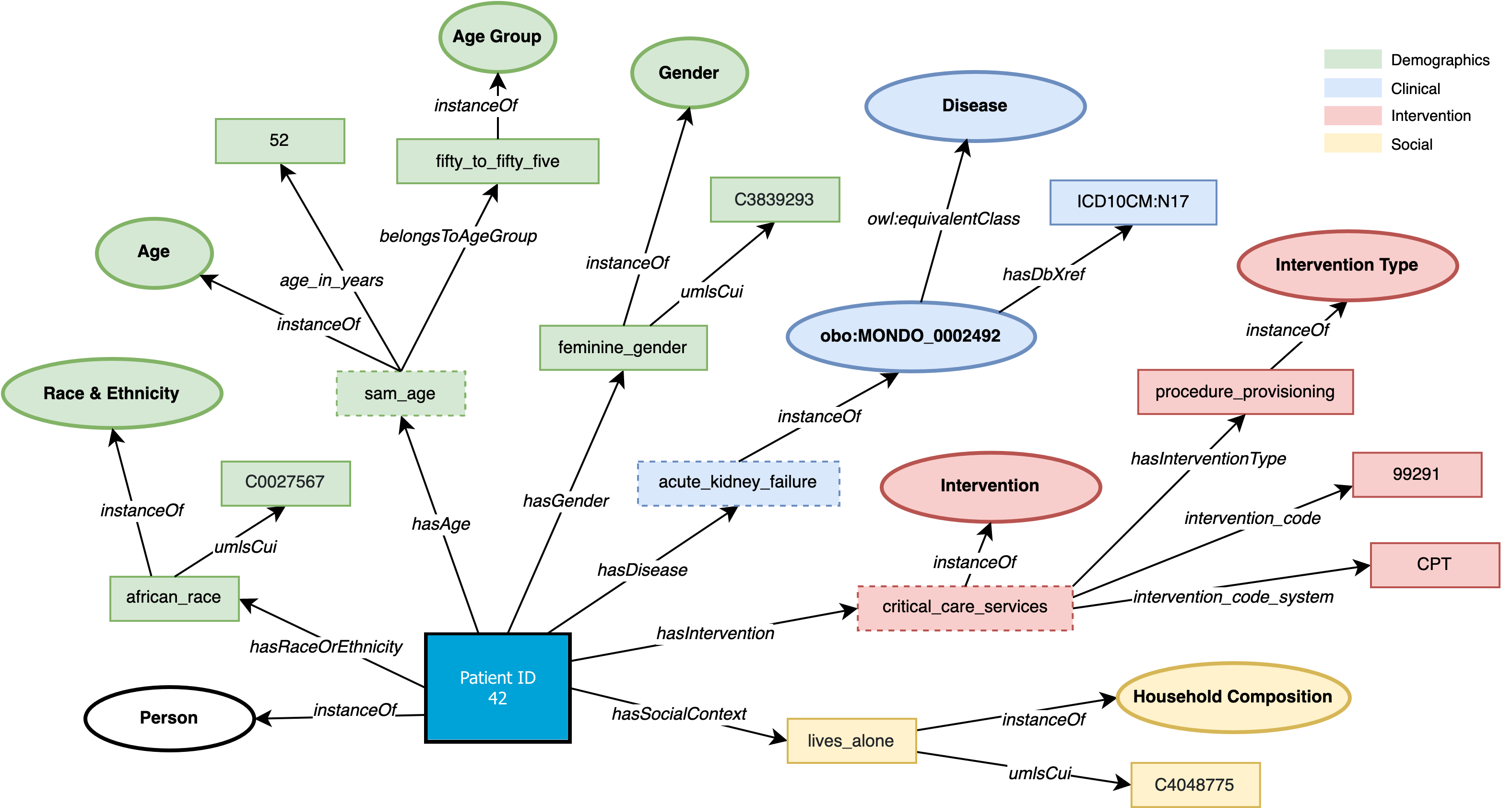}
  \vspace{-4.9mm}
   \caption{An example Person-Centric Knowledge: The graph represents the demographics, clinical, intervention, and social aspects of the patient.}
   \vspace{-7.5mm}
\end{figure}

The HSPO ontology provides the knowledge schema  used to create the PKGs. A PKG is extracted for every admission record of the processed data. The ontology represents the different classes (e.g. Religion, Disease, etc.), instances/individuals of the classes (e.g. Acute Kidney Failure), relations between the classes (e.g. hasDisease, hasSocialContext, etc.), and data properties (e.g. the instances of the class Age has a data property \textit{age\_in\_years} that is an integer number). We use the rdflib python library for the implementation. The extracted knowledge graphs follow the Resource Description Framework (RDF) \cite{decker2000semantic} format (Fig. 3), which is the primary foundation for the Semantic Web. Thus, SPARQL Query Language \cite{cyganiak2005relational,perez2006semantics} can be used to query, search and perform operations on the graphs. 

\vspace{-5mm}

\section{Evaluation}



\vspace{-3mm}

The evaluation section reflects on the validity, reliability, and effectiveness of the person-centric graphs in downstream tasks. We evaluate the patient representation learning using person-centric graphs and a GNN on a specific ICU readmission prediction task. We provide insights into the applicability, benefits, and challenges of the proposed solution.

\vspace{-4mm}

\subsection{Data Transformation}

\vspace{-2.5mm}

The extracted graphs in RDF format cannot be used to train GNNs. Hence, a transformation step is implemented to transform graphs into the format (.dt files) of PyTorch Geometric \cite{fey2019fast}, as we adopt this framework to build the models. The transformed graphs consist of the initial representations (initialized embeddings) of the nodes and the adjacency matrices for each relation type. They follow a triplet-based format, where the triplets have the following format: \textit{[entity 1, relation type, entity 2]} (e.g. [patient, hasDisease, acute renal failure]).

Training GNNs using person-centric graphs as input is a relatively unexplored field and defining \textit{a priori} the most useful graph structure is not a trivial task. Hence, we experiment with 4 different graph versions (Fig. 4) to find the most suitable structure for the downstream task given the available data. The strategy to define the different graph structures is described as follows: starting from a detailed version, we progressively simplify the graph by reducing the heterogeneity with relation grouping. We highlight that finding the optimal level of heterogeneity is an open and challenging research question as it depends on the available data, the downstream task, and the model architecture.

More precisely, the first version of the graph structure is aligned with the schema provided by the HSPO ontology and includes 8 relation types (\textit{hasDisease}, \textit{hasIntervention}, \textit{hasSocialContext}, \textit{hasRaceOrEthnicity}, \textit{followsReligion}, \textit{hasGender}, \textit{hasMaritalStatus}, and \textit{hasAge}). The detailed demographic relations are grouped under the \textit{hasDemographics} relation in the second version. The third version is the most simplified containing only the \textit{has} relation type.

Lastly, we present the fourth version of the graph with the inclusion of group nodes to explore the effectiveness of graph expressivity on learnt representation. The design of the fourth version is based on the assumption that the summarization of the corresponding detailed information (e.g. the disease information is summarized in the representation of the \textit{Disease} node) and the learning of a grouped representation during training can be beneficial for the performance of the models. To explore this assumption in the experimental setup, we introduce a node type \textit{group node} and add 7 group nodes (\textit{Diseases}, \textit{Social Context}, \textit{Demographics}, \textit{Age}, \textit{Interventions}, \textit{Procedures}, and \textit{Medication}). For the inclusion of the group nodes in the graph, we introduce the general relation type \textit{has}. Fig. 4 presents the four directed graph structures. We also include the undirected corresponding versions in the experiments.

The nodes of each graph are initialized using the bag-of-words (BOW) approach. Hence, the initial representations are sparse. We create the vocabulary using the textual descriptions of the end nodes (diagnoses names, medication, etc.) and the description of the group nodes (patient, social context, diseases, interventions, medication, procedures, demographics, and age). The final vocabulary consists of 3,723 nodes. Alternatively to the introduction of sparsity using the BOW approach, a language model, such as BioBERT \cite{lee2020biobert}, PubMedBERT \cite{gu2021domain}, and CharacterBERT \cite{el-boukkouri-etal-2020-characterbert,theodoropoulos-etal-2021-imposing}, can be used for the node initialization. We provide this capability in our open-source framework.

\begin{figure}[!h]
  \vspace{-7mm}
  \centering
  \includegraphics[width=0.95\textwidth]{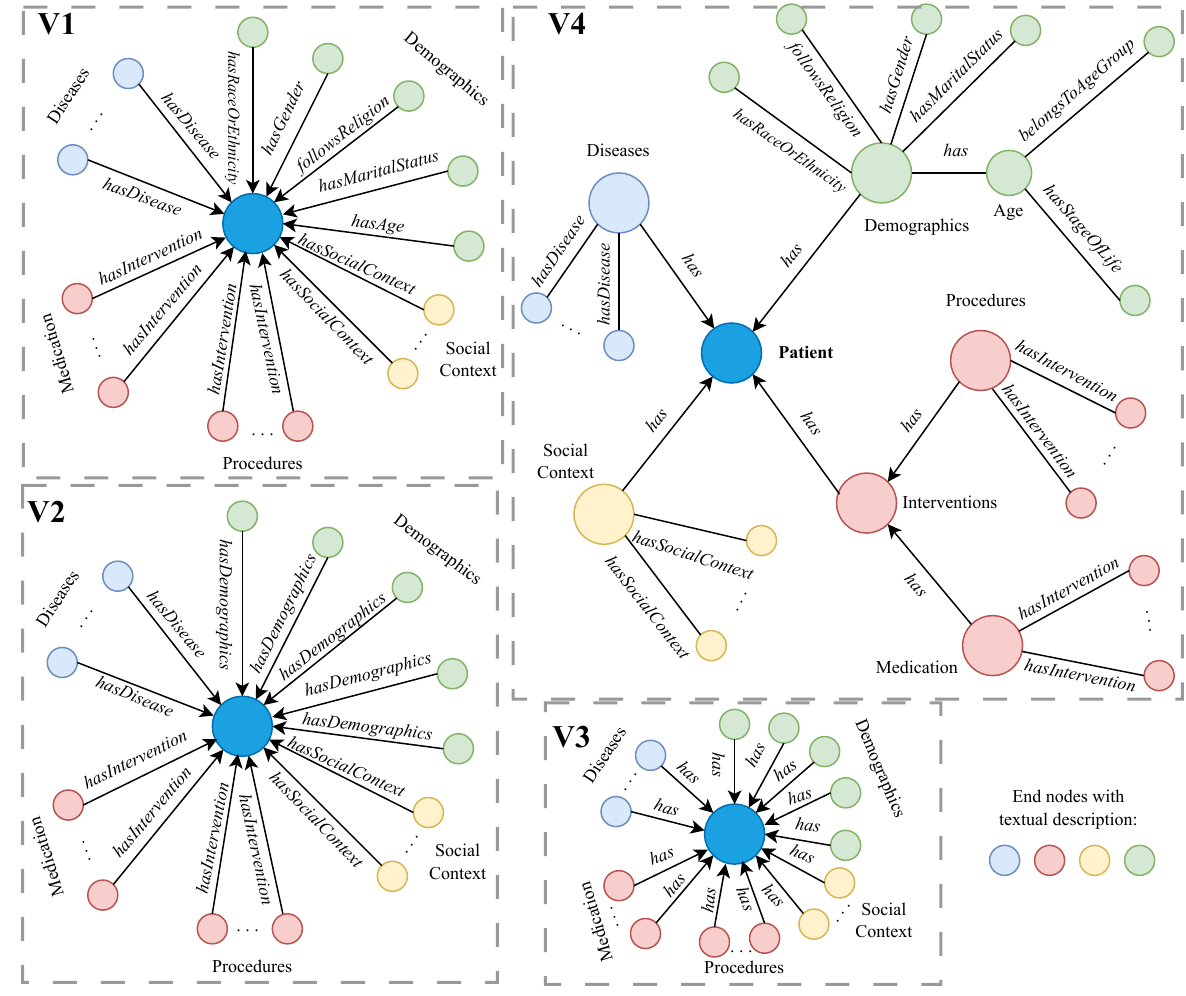}
  \vspace{-5.5mm}
   \caption{The 4 directed graph structures of the study are presented. The colors of the nodes represent the different types of information and are consistent across the versions.}
   \vspace{-11mm}
\end{figure}

\subsection{Graph Neural Networks and Baseline Models}

\vspace{-2mm}

We experiment with two different GNN architectures and each of them has two variations. The difference between the variations lies in the final layer of the model which is convolutional or linear (Fig. 5).  The first model (PKGSage) is based on Sage Graph Convolution Network \cite{hamilton2017inductive} and the second (PKGA) utilizes the Graph Attention Network (GAT) architecture \cite{velivckovic2017graph,brody2021attentive}. Given a graph structure \textit{G} with \textit{N} nodes, a sequence of transformations \textit{T} is applied, and the final prediction \textit{p} of the model is extracted as follows:

\vspace{-4mm}

\begin{equation}
    X_{k, i} = \sigma(T_{i}(X_{k, i-1}, G)), with \, k \in [1, ..., N] \, and \, i \in [1, 2, 3],
\end{equation}

\vspace{-5.5mm}

\begin{equation}
    p = \sigma(X_{n, 3}),
\end{equation}

where $T_{i}$ is the transformation (Sage Convolution, GAT Convolution, or linear) of the $i^{th}$ layer, $X_{k, i}$ is the $k$ node representation after the $T_{i}$ transformation, $X_{n, 3}$ is the final output of the last layer for the \textit{patient} node, and $\sigma$ is the activation function. ReLU is used as the activation function for the first two layers and Sigmoid for the last layer. In principle, the graphs are multi-relational and this can lead to rapid growth in the number of model parameters. To address this issue, we apply basis decomposition \cite{schlichtkrull2018modeling} for weight regularization. 

We incorporate a set of baseline classifiers to compare the graph-based approach with traditional machine-learning algorithms. Particularly, k-nearest neighbors (KNN) (k is set to 5 based on the average performance in the validation set), linear and non-linear Support Vector Machines (L-SVM and RBF-SVM respectively) \cite{cortes1995support}, Decision Tree (DT) \cite{breiman2017classification}, AdaBoost \cite{freund1997decision,hastie2009multi}, Gaussian Naive Bayes (NB), and Gaussian Process (GP) \cite{williams2006gaussian} with radial basis function (RBF) kernel are included in the study. We apply one-hot encoding to transform the textual descriptions into numerical features for the diagnoses, medication, and procedures. The remaining features (gender, religion, marital status, age group, race, employment, housing conditions, and household composition) are categorical, so we encode each category using mapping
to an integer number. For a fair comparison, feature engineering or feature selection techniques are not implemented, since the graph-based models include all the available information extracted from the EHRs. 

\begin{figure}[!h]
  \vspace{-6mm}
  \centering
  \includegraphics[width=\textwidth]{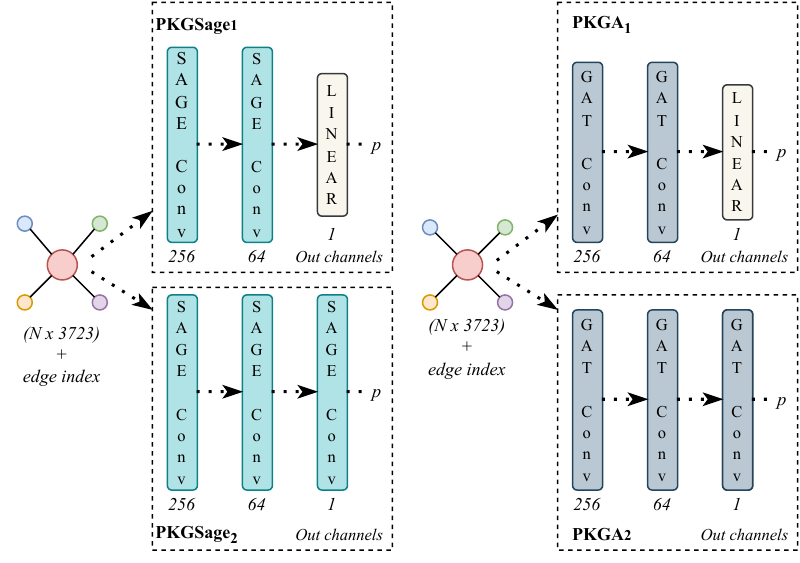}
  \vspace{-9.5mm}
   \caption{The input of the model is a graph with \textit{N} number of nodes and edge index per relation type. The dimension of the initial representation of the nodes is 3,723. The PKGSage models leverage the Sage Convolution as the main transformation step while the PKGA models include the GAT Convolution module. The final output of the models is the readmission probability of the patient.}
   \vspace{-8mm}
\end{figure}

\vspace{-3mm}

\subsection{Experimental Setup}

\vspace{-2.5mm}

We create 10 different balanced dataset splits for experimentation to overcome the imbalance problem. In detail, the 14,113 non-readmission cases are randomly divided into 10 folds. Combined with the 1,428 readmission cases, these folds constitute the final 10 balanced dataset splits. For each balanced split, 5-fold cross-validation is applied. We highlight that we use the same splits across the different experimental settings to have a fair comparison. The Adam optimizer \cite{kingma2014adam} is used and the learning rate is set to 0.001. The models are trained for 100 epochs and the best model weights are stored based on the performance in the validation set (15\% of the training set). The number of bases, for the basis decomposition method \cite{schlichtkrull2018modeling}, is 3, and the batch size is set to 32. 

\vspace{-3mm}

\subsection{Results}

\vspace{-3mm}

In this subsection, we present the average results, across the different runs and folds of 5-fold cross-validation, of the models using different graph versions (Tab. 2) and compare the performance of the best models with various machine learning classifiers (Tab. 3). Starting with the intra-model comparison, the PKGSage model performs best using the first directed graph version with 62.16\% accuracy and the third undirected graph version with 68.06\% F1-score. The PKGA model achieves 61.69\% accuracy and 67.49\% F1-score using the third undirected graph version. Overall, the end-to-end convolution strategy is advantageous as the second variation of the models (PKGSage\textsubscript{2}, PKGA\textsubscript{2}) performs better than the first variation with the linear final layer in most cases. Nonetheless, the best performance is achieved by the PKGSage\textsubscript{1} model. The inter-model comparison reveals that both models (PKGSage, PKGA) achieve similar results. 

\begin{table}[!b]
  \vspace{-6mm}
  \centering
  \caption{Results: Performance of the models with different graph versions.}
  \vspace{-5mm}
  \resizebox{\columnwidth}{!}{
  \begin{threeparttable}
  \label{tab:freq}
  \begin{tabular}{cccccc|cccccc}
        \textbf{G.}\tnote{1} & \textbf{D.}\tnote{2} & \textbf{Model} & \textbf{Accuracy} & \textbf{F1-Score} & & & \textbf{G.}\tnote{1} & \textbf{D.}\tnote{2} & \textbf{Model} & \textbf{Accuracy} & \textbf{F1-Score}\\
        \hline
        \multirow{8}{*}{V1} & \parbox[t]{2mm}{\multirow{4}{*}{\rotatebox[origin=c]{90}{Undirected}}} & PKGSage\textsubscript{1} & \textit{61.72} \textpm{ 0.98} & \textit{67.86} \textpm{ 1.31} & & & \multirow{8}{*}{V3} & \parbox[t]{2mm}{\multirow{4}{*}{\rotatebox[origin=c]{90}{Undirected}}} & PKGSage\textsubscript{1} & \textit{61.69} \textpm{ 0.91} & \textbf{68.06} \textpm{ 1.14}\\
                                                        & & PKGSage\textsubscript{2} & 61.27 \textpm{ 0.92} & 67.44 \textpm{ 1.34} & & & 
                                                        & & PKGSage\textsubscript{2} & 61.48 \textpm{ 0.93} & 67.5 \textpm{ 1.21}\\
                                                        & & PKGA\textsubscript{1} & 59.73 \textpm{ 1.08} & 66.32 \textpm{ 1.12} & & &                 & & PKGA\textsubscript{1} & \textit{61.69} \textpm{ 0.71} & 66.32 \textpm{ 1.56}\\
                                                        & & PKGA\textsubscript{2} & 61.39 \textpm{ 0.89} & 66.58 \textpm{ 1.56} & & &                 & & PKGA\textsubscript{2} & 61.46 \textpm{ 0.82} & 67.49 \textpm{ 1.76}\\
        \cmidrule{2-6}
        \cmidrule{9-12}
                            & \parbox[t]{2mm}{\multirow{4}{*}{\rotatebox[origin=c]{90}{Directed}}} & PKGSage\textsubscript{1} & \textbf{62.16} \textpm{ 0.89} & \textit{67.93} \textpm{ 1.35} & & & & \parbox[t]{2mm}{\multirow{4}{*}{\rotatebox[origin=c]{90}{Directed}}} & PKGSage\textsubscript{1} & \textit{50.2} \textpm{ 0.51} & \textit{64.21} \textpm{ 1.74}\\
                                                        & & PKGSage\textsubscript{2} & 61.72 \textpm{ 0.73} & 67.2 \textpm{ 1.68} & & &               & & PKGSage\textsubscript{2} & 49.93 \textpm{ 0.57} & 61.43 \textpm{ 1.83}\\
                                                        & & PKGA\textsubscript{1} & 60.5 \textpm{ 1.26} & 66.52 \textpm{ 1.51} & & &                  & & PKGA\textsubscript{1} & \textit{50.2} \textpm{ 0.63} & 61.54 \textpm{ 1.91}\\
                                                        & & PKGA\textsubscript{2} & 61.38 \textpm{ 0.87} & 67 \textpm{ 1.65} & & &          
                                                        & & PKGA\textsubscript{2} & 49.98 \textpm{ 0.61} & 61.44 \textpm{ 1.87}\\  
        \hline
        \multirow{8}{*}{V2} & \parbox[t]{2mm}{\multirow{4}{*}{\rotatebox[origin=c]{90}{Undirected}}} & PKGSage\textsubscript{1} & 60.43 \textpm{ 0.63} & \textit{67.85} \textpm{ 1.22} & & & \multirow{8}{*}{V4} & \parbox[t]{2mm}{\multirow{4}{*}{\rotatebox[origin=c]{90}{Undirected}}}& PKGSage\textsubscript{1} & 49.5 \textpm{ 0.55} & \textit{59.77} \textpm{ 2.54}\\
                                                        & & PKGSage\textsubscript{2} & \textit{60.93} \textpm{ 0.81} & 67.06 \textpm{ 1.45} & & &     & & PKGSage\textsubscript{2} & 54.5 \textpm{ 1.4} & 59.51 \textpm{ 2.36}\\
                                                        & & PKGA\textsubscript{1} & 58.95 \textpm{ 0.78} & 66.21 \textpm{ 1.36} & & &                 & & PKGA\textsubscript{1} & 52.08 \textpm{ 1.68} & 53.1 \textpm{ 1.88}\\
                                                        & & PKGA\textsubscript{2} & 60.24 \textpm{ 0.99} & 67.37 \textpm{ 1.31} & & &                 & & PKGA\textsubscript{2} & \textit{57.9} \textpm{ 0.91} & 58.67 \textpm{ 0.9}\\
        \cmidrule{2-6}
        \cmidrule{9-12}
                            & \parbox[t]{2mm}{\multirow{4}{*}{\rotatebox[origin=c]{90}{Directed}}} & PKGSage\textsubscript{1} & 51.26 \textpm{ 0.81} & \textit{65.1} \textpm{ 1.58} & & & & \parbox[t]{2mm}{\multirow{4}{*}{\rotatebox[origin=c]{90}{Directed}}}& PKGSage\textsubscript{1} & 49.45 \textpm{ 0.58} & 58.41 \textpm{ 2.75}\\
                                                        & & PKGSage\textsubscript{2} & \textit{51.61} \textpm{ 0.76} & 61.75 \textpm{ 1.26} & & &     & & PKGSage\textsubscript{2} & 54.79 \textpm{ 1.79} & 59.87 \textpm{ 2.67}\\
                                                        & & PKGA\textsubscript{1} & 51 \textpm{ 0.81} & 64.86 \textpm{ 1.27} & & &                    & & PKGA\textsubscript{1} & 51.01 \textpm{ 1.29} & \textit{62.92} \textpm{ 1.54}\\
                                                        & & PKGA\textsubscript{2} & 51.3 \textpm{ 0.91} & 63.84 \textpm{ 1.43} & & &                  & & PKGA\textsubscript{2} & \textit{57.32} \textpm{ 1.54} & 60.16 \textpm{ 2.28}\\ 
        \hline
    \end{tabular}
  \begin{tablenotes}
      \item [1] G.: Graph
      \item [2] D.: Direction
  \end{tablenotes}
  \end{threeparttable}}
  \vspace{-6mm}
\end{table}

The graph's structure is essential since the performance varies according to the graph version. More precisely, the results are comparable with less notable differences using the undirected/directed first version of the graph, and the undirected second and third versions. However, the performance degradation is significant when the directed second and third versions are utilized. In both versions, we observe that all links point to the central \textit{Patient} node (Fig. 4), and the representations of the remaining nodes are not updated during training, imposing a challenge on the trainability of the models. However, the same structure is present in the directed first version where no performance drop is noticed. Another possible reason is the level of heterogeneity of the graph. The first version has 8 relation types, while the second and third versions contain 4 and 1 relation types respectively. Given these, we conclude that the direction of the links and graph heterogeneity are crucial for the final performance of the models. Leveraging the fourth graph version is not advantageous since the models achieve worse performance. This indicates that introducing the group nodes and additional expressibility can be an obstacle to the trainability and performance of the models. We observe saturation and stability problems \cite{pmlr-v9-glorot10a} during training when the fourth version is used.

The best-performing models of the study also significantly outperform the baseline models. We notice an improvement of 3.62\% in F1-Score. Based on the accuracy metric, only the SVM classifiers (linear and non-linear) achieve comparable results. The results illustrate the potential of the PKG-based models in downstream predictive tasks and particularly in ICU readmission prediction. 

\begin{table}[!h]
    \vspace{-5mm}
    \caption{Results: Comparison with baseline models}
    \vspace{-3mm}
    \centering
    \begin{tabular}{ccccccccc}
        \textbf{Metric} & & \textbf{DT} & & \textbf{AdaBoost} & & \textbf{NB} & & \textbf{GP}\\
        \hline
        Accuracy & & 55.34 \textpm{ 0.55} & & 60.03 \textpm{ 0.65} & & 53.92 \textpm{ 1.44} & & 56.82 \textpm{ 0.57} \\
        \hline
        F1-Score & & 57.5 \textpm{ 0.65} & & 59.01 \textpm{ 1.44} & & 39.93 \textpm{ 3.56} & & 52.53 \textpm{ 1.62} \\
        \hline
        \hline
        \textbf{Metric} & & \textbf{KNN} & & \textbf{L-SVM} & & \textbf{RBF-SVM} & & \textbf{PKGSage\textsubscript{1}}\\
        \hline
        Accuracy & & 57.31 \textpm{ 0.54} & & 61.58 \textpm{ 0.62} & & 62.11 \textpm{ 0.65} & & \textbf{62.16} \textpm{ 0.89}\\
        \hline
        F1-Score & & 50.9 \textpm{ 1.73} & & 61.45 \textpm{ 1.49} & & 64.44 \textpm{ 1.49} & & \textbf{68.06} \textpm{ 1.14} \\
        \hline
    \end{tabular}
    \vspace{-12mm}
\end{table}


\subsection{Ablation Study}

\vspace{-2.5mm}

Following the observation that the data is incomplete (Tab. 1), we conduct an ablation study to probe the robustness of our approach in handling missing information. The following hypotheses are drawn: 

\begin{itemize}
    \vspace{-1.5mm}
    \item The pure clinical view of the patient (medication, diseases, and procedures) is very important in predicting ICU readmission. (H1)
    \item The exclusion of additional information results in lower performance. (H2)
    \vspace{-1.5mm}
\end{itemize}

We apply the ablation study using the best-performing model PKGSage\textsubscript{1} (Tab. 3) with the undirected third graph version (PKGSage\textsubscript{1}UnV\textsubscript{3}) and the directed first graph version (PKGSage\textsubscript{1}DV\textsubscript{1}). In the first step, we exclude one facet of the data (e.g. medication, procedures) and evaluate performance degradation. To address the H2 hypothesis and to further test model robustness, in the next step, we exclude two facets in a complete manner (as depicted in Tab. 4). 


\begin{table}[!h]
    \vspace{-3mm}
    \caption{Ablation Study}
    \vspace{-2mm}
    \centering
    \resizebox{\textwidth}{!}{
    \begin{tabular}{ccccc}
        \multirow{2}{*}{\textbf{Excluded Information}} & \multicolumn{2}{c|}{\textbf{PKGSage\textsubscript{1} undirected V3}} & \multicolumn{2}{c}{\textbf{PKGSage\textsubscript{1} directed V1}} \\
        \cmidrule{2-5}
        & \textbf{Accuracy} & \textbf{F1-Score} & \textbf{Accuracy} & \textbf{F1-Score}\\
        \hline
        - & 61.69 & 68.06 & 62.16 & 67.93\\
        \hline
        Social aspect & 60.68 ($\downarrow1.01$) & 67.63 ($\downarrow0.43$) & 60.5 ($\downarrow1.66$) & 67.64 ($\downarrow0.29$)\\
        \hline
        Medication & 59.7 ($\downarrow1.99$) & 66.98 ($\downarrow1.08$) & 60.03 ($\downarrow2.13$) & \textbf{66.8} ($\downarrow1.13$)\\
        \hline
        Procedures & 60.42 ($\downarrow1.27$) & 67.43 ($\downarrow0.63$) & 59.97 ($\downarrow2.19$) & 67.56 ($\downarrow0.37$)\\
        \hline
        Diseases & \textbf{59.69} ($\downarrow2$) & \textbf{66.87} ($\downarrow1.19$) & \textbf{59.59} ($\downarrow2.57$) & 67.48 ($\downarrow0.45$)\\
        \hline
        Demographics & 60.43 ($\downarrow1.26$) & 67.43 ($\downarrow0.63$) & 60.16 ($\downarrow2$) & 67.54 ($\downarrow0.39$)\\
        \hline
        \hline
        Social aspect and Medication & 60.13 ($\downarrow1.56$) & 66.84 ($\downarrow1.22$) & 60.01 ($\downarrow2.15$) & 66.67 ($\downarrow1.26$)\\
        \hline
        Social aspect and Procedures & 60.13 ($\downarrow1.56$) & 67.99 ($\downarrow0.07$) & 59.87 ($\downarrow2.29$) & 67.34 ($\downarrow0.59$)\\
        \hline
        Social aspect and Diseases & 59.87 ($\downarrow1.82$) & 66.62 ($\downarrow1.44$) & 59.6 ($\downarrow2.56$) & 67.18 ($\downarrow0.75$)\\
        \hline
        Social aspect and Demographics & 60.64 ($\downarrow1.05$) & 66.84 ($\downarrow1.22$) & 60.14 ($\downarrow2.02$) & 66.88 ($\downarrow1.05$)\\
        \hline
        Medication and Procedures & \textbf{58.76} ($\downarrow2.93$) & 66.59 ($\downarrow1.47$) & 59.47 ($\downarrow2.69$) & 66.86 ($\downarrow1.07$)\\
        \hline
        Medication and Diseases & 59.49 ($\downarrow2.2$) & \textbf{65.69} ($\downarrow2.37$) & 59.72 ($\downarrow2.44$) & \textbf{66.06} ($\downarrow1.87$)\\
        \hline
        Medication and Demographics & 59.84 ($\downarrow1.85$) & 66.8 ($\downarrow1.26$) & 59.9 ($\downarrow2.26$) & 66.56 ($\downarrow1.37$)\\
        \hline
        Procedures and Diseases & 59.32 ($\downarrow2.37$) & 66.62 ($\downarrow1.44$) & \textbf{58.53} ($\downarrow3.63$) & 66.38 ($\downarrow1.55$)\\
        \hline
        Procedures and Demographics & 60.63 ($\downarrow1.06$) & 67.69 ($\downarrow0.37$) & 60.01 ($\downarrow2.15$) & 67.67 ($\downarrow0.26$)\\
        \hline
        Diseases and Demographic & 59.53 ($\downarrow2.16$) & 66.85 ($\downarrow1.21$) & 59.56 ($\downarrow2.6$) & 66.52 ($\downarrow1.73$)\\
        \hline
    \end{tabular}}
    \vspace{-5mm}
\end{table}

The exclusion of the disease information results in the most significant performance decline in the accuracy for both model versions and in the F1-score for the PKGSage\textsubscript{1}UnV\textsubscript{3} model (Tab. 4). Excluding the medication information leads to 1.13\% drop in F1-score for the PKGSage\textsubscript{1}DV\textsubscript{1} model. The results of the ablation study only partially support the H1 hypothesis. The robustness of the models in handling missing information is profound as the performance deterioration is limited. In the worst case, accuracy and F1-score drop by 2.57\% and 1.13\% respectively. A similar pattern is revealed when we remove two facets of the data as the removal of clinical information is reflected in lower performance. More precisely, the exclusion of the medication and disease information leads to 2.37\% and 1.87\% drop in F1-score for the PKGSage\textsubscript{1}UnV\textsubscript{3} and PKGSage\textsubscript{1}DV\textsubscript{1} models correspondingly. PKGSage\textsubscript{1}UnV\textsubscript{3} is less accurate by 2.93\% when medication and procedures nodes are absent, and PKGSage\textsubscript{1}DV\textsubscript{1} achieves 58.53\% accuracy (3.63\% reduction) when procedure and disease data are excluded. Overall, the performance of the models is robust even when two out of three clinical facets (medication, diseases, and procedures) are unavailable. We highlight that stability and robustness are key properties, especially in healthcare data where missing information is inevitable due to privacy issues, system or human errors.

\vspace{-3.5mm}

\section{Discussion} 

\vspace{-3.5mm}

The proposed end-to-end framework for person or entity-centric KGs has impacts on both the technology and industry and we highlight the benefits and challenges of adopting KG technologies in practice. We focus on the healthcare domain yet present a solution that is, by design, generalisable across domains and it is also not restricted by the final predictive task. The proposed approach was evaluated using complex real-world data, which is imbalanced, heterogeneous, and sparse, and the ontology efforts and approach were reviewed by domain experts from multiple organisations. Indeed, planned next steps include a new use case in inflammatory bowel disease (IBD) which can be studied across institutions. IBD is a compelling next use case not only because a significant amount of the information needed by clinicians is buried in unstructured text notes but also because multiple facets, including diet, behaviors, race, and ethnicity or environmental factors are all thought to contribute to disease progression and outcomes \cite{lloyd2019multi,liu2022}. The proposed star-shaped KG approach and ontology will not only allow data from disparate sources and types to be meaningfully combined, as already demonstrated, but also allow pertinent new research questions to be addressed. This paves the way towards a comprehensive and holistic 360\textdegree view of the person to be created from multiple data sources and could be further generalised to patient cohorts or groups of individuals.

This approach is also scalable as multiple PKGs can be created, one for each patient, unlike traditional approaches relying on very large general knowledge KGs or others which have reported scalability issues \cite{schlichtkrull2018modeling}. The topology of traditional KGs used to query and infer knowledge might not be easily applied to learn graph representations using GNNs primarily due to their large size. Our experiments show that reducing the heterogeneity with relation grouping has an effect on F1-score provided that the PKG structure is generally fit for learning patient representations using GNNs. We also observe that the proposed framework is able to generalise even with limited amount of data.

Furthermore, the ontology design process can be used as guidance for the creation of other entity-centric ontologies and the HSPO is continuing to be expanded with new domains and facets. The open-sourced implementation for the PKG creation can be reused, with necessary adjustments, to extract entity-centric knowledge graphs in RDF or PyTorch Geometric applicable format. Finally, a wide range of predictive tasks using neural networks, or inferences using the KGs produced can be addressed, even if these will be constrained by the availability and expressiveness of the data and provided annotations. 

We highlight the applicability of the framework through a readmission prediction task using the PKGs and GNN architectures. Following the proposed paradigm, other classification tasks, such as mortality prediction and clinical trial selection, can be undertaken. The efficient patient representation may also be leveraged for clustering and similarity grouping applications. 

\vspace{-3.5mm}

\section{Conclusion and Future Work}

\vspace{-3.5mm}

This paper proposes a new end-to-end representation learning framework to extract entity-centric ontology-driven KGs from structured and unstructured data. Throughout this paper, we describe how KG technologies can be used in combination with other technologies and techniques (learning entity representation using GNNs, predicting outcomes (prediction tasks) using neural networks and learnt representation) to drive practical industry applications with an example in the healthcare industry. The open-sourced framework and approach are scalable and show stability and robustness when applied to complex real-world healthcare data. We plan to extend and apply the PKG and the proposed framework to new use cases which will further drive adoption. In healthcare, we envisage that this work can unlock new ways of studying complex disease patterns and drive a better understanding of disease across population groups. 

\vspace{-3mm}

\paragraph*{Supplemental Material Availability:} Source code is available in the official GitHub repository: https://github.com/IBM/hspo-ontology

\vspace{-3mm}

\section*{Acknowledgements}

\vspace{-4mm}

We would like to acknowledge the teams from Cleveland Clinic (Dr. Thaddeus Stappenbeck, Dr. Tesfaye Yadete) and Morehouse School of Medicine (Prof. Julia Liu, Dr. Kingsley Njoku) and colleagues from IBM Research (Vanessa Lopez, Marco Sbodio, Viba Anand, Elieen Koski) for their support and insights.  
%
%
%
\bibliographystyle{splncs04}
\bibliography{mybibliography}

\end{document}